\documentclass[runningheads]{llncs}

\usepackage{eccv}

\usepackage{eccvabbrv}

\usepackage{graphicx}
\usepackage{booktabs}
\usepackage{enumitem}
\usepackage{multirow}
\usepackage{wrapfig}
\usepackage{bbm}
\def\etal{\emph{et al.}}
\usepackage[accsupp]{axessibility}  % Improves PDF readability for those with disabilities.

\usepackage[pagebackref,breaklinks,colorlinks,citecolor=eccvblue]{hyperref}
\usepackage{orcidlink}

\begin{document}

% ---------------------------------------------------------------
% TODO REVIEW: Replace with your title
\title{UniCode \includegraphics[height=12pt]{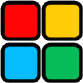} : Learning a Unified Codebook for Multimodal Large Language Models} 

% TODO REVIEW: If the paper title is too long for the running head, you can set
% an abbreviated paper title here. If not, comment out.
\titlerunning{Abbreviated paper title}

% TODO FINAL: Replace with your author list. 
% Include the authors' OCRID for the camera-ready version, if at all possible.
%\author{First Author\inst{1}\orcidlink{0000-1111-2222-3333} \and
%Second Author\inst{2,3}\orcidlink{1111-2222-3333-4444} \and
%Third Author\inst{3}\orcidlink{2222--3333-4444-5555}}

\author{
Sipeng Zheng$^{1}$, Bohan Zhou$^{2}$, Yicheng Feng$^{2}$, Ye Wang$^{1}$,
Zongqing Lu$^{1,2}$\thanks{Zongqing Lu is the corresponding author.}  \\
$^{1}$Beijing Academy of Artificial Intelligence (BAAI)
$^{2}$School of Computer Science, Peking University
\\
{\tt\small spzheng@baai.ac.cn }
{\tt\small zhoubh@stu.pku.edu.cn }
{\tt\small yewang@stu.ecnu.edu.cn }
{\tt\small\{fyc813, zongqing.lu\}@pku.edu.cn }
}
\institute{}
%\authorrunning{S Zheng et al.}

%\email{\{abc,lncs\}@uni-heidelberg.de}}

\maketitle

\begin{abstract}
In this paper, we propose \textbf{UniCode}, a novel approach within the domain of multimodal large language models (MLLMs) that learns a unified codebook to efficiently tokenize visual, text, and potentially other types of signals. 
This innovation addresses a critical limitation in existing MLLMs: their reliance on a text-only codebook, which restricts MLLM's ability to generate images and texts in a multimodal context.
Towards this end, 
we propose a language-driven iterative training paradigm, coupled with an in-context pre-training task we term ``image decompression'', enabling our model to interpret compressed visual data and generate high-quality images.
The unified codebook empowers our model to extend visual instruction tuning to non-linguistic generation tasks.
Moreover, UniCode is adaptable to diverse stacked quantization approaches in order to compress visual signals into a more compact token representation.
Despite using significantly fewer parameters and less data during training, Unicode demonstrates promising capabilities in visual reconstruction and generation.
It also achieves performances comparable to leading MLLMs across a spectrum of VQA benchmarks.
\end{abstract}

\section{Introduction}
\label{sec:intro}

\begin{figure}
\centering
\includegraphics[width=0.85\textwidth]{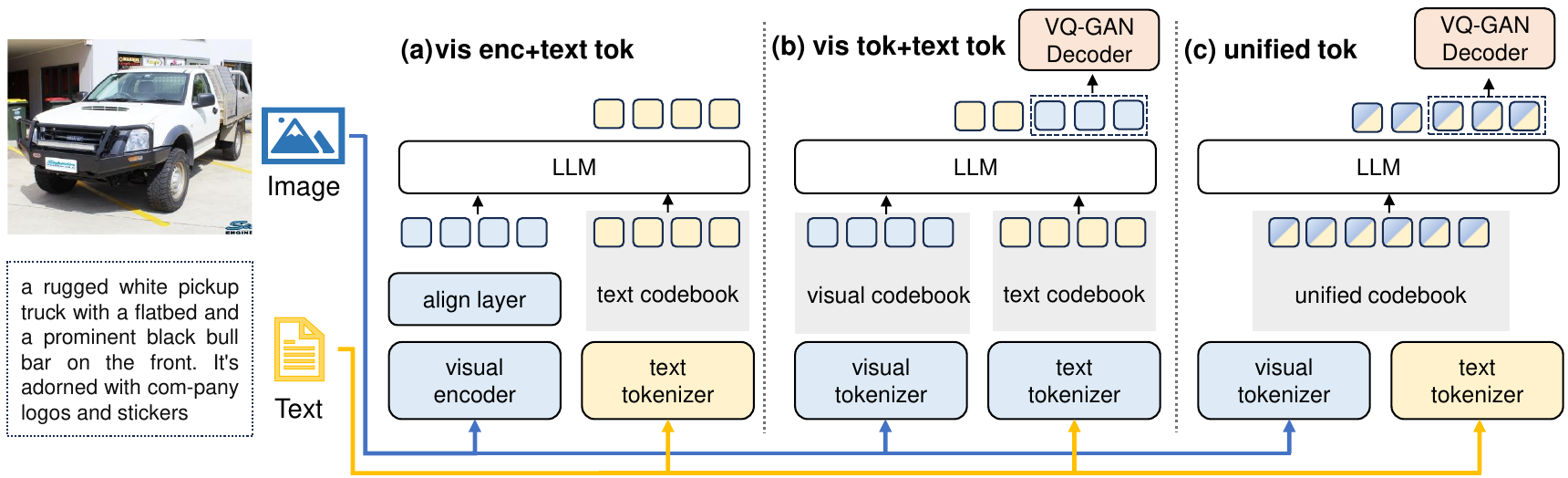}
\caption{
Three paradigms of MLLMs: \textbf{(a) vis enc+text tok} incorporates a lightweight module to align visual signals with the LLM, specifically designed for languge generation;
\textbf{(b) vis tok+text tok} concatenates the text codebook with quantized visual tokens, significantly increasing the computational cost and complexity;
\textbf{(c) unified tok} learns a unified codebook to interpret both visual and text modalities without additional modules. 
We explore the last option by proposing UniCode in this work.
}
\label{fig:intro} 
\vspace{-0.5cm}
\end{figure}

The rapid development of large language models (LLMs)~\cite{chatgpt, touvron2023llama, touvron2023llama2} has spurred growing interest in their multimodal counterparts~\cite{alayrac2022flamingo, li2023blip, liu2023visual}.
Empowered by LLMs, existing foundation models have shown remarkable capabilities in multimodal understanding, spanning from basic image classification~\cite{radford2021learning,yuan2021florence}
and captioning~\cite{wang2022git,li2023blip}, to more intricate tasks such as making strategic high-level plans for open-world agents~\cite{zheng2023steve,fengllama}.
As illustrated in Figure~\ref{fig:intro} (a), these works rely on a lightweight module such as a multimodal projector~\cite{liu2023visual, li2023blip}
to seamlessly map visual signals into LLM's textual space with minimal training cost.

% Use figure* for multi-column figure

Progress has been made though, most multimodal large language models (MLLMs) are still limited to language generation.
This limitation stems from their reliance on text-only codebooks, which restricts their application across diverse scenarios, such as image generation~\cite{lee2022autoregressive}.
Note that images, like text, can be tokenized into a series of discrete codes through Vector Quantization (VQ)~\cite{van2017neural,esser2021taming,lee2022autoregressive}, which suggests a straightforward strategy to enhance MLLMs: extending the LLM's codebook to include visual codes~\cite{lu2023unified, cui2023efficient} as shown in Figure~\ref{fig:intro} (b).
However, this approach introduces new challenges.
Firstly, it requires considerable effort to overcome the substantial modality gap between visual and text codes. 
Secondly, enlarging the codebook leads to an upsurge in model parameters and risks ``codebook collapse''~\cite{dhariwal2020jukebox}, where the model overly relies on a limited set of codes, posing significant obstacles in the training of MLLMs.

Instead of expanding the codebook size, we pose a question: ``\textit{Is it feasible to learn a unified codebook capable of quantizing language, vision, and potentially other modalities?}''
As illustrated in Figure~\ref{fig:intro} (c), a unified codebook could seamlessly integrade various data types, thereby equipping MLLMs with the ability to generate non-linguistic content without the need for additional parameters or specialized modules. 
Recent initiatives have explored to map  visual signals into the text token space of a frozen LLM.
Yet, these efforts offen yield inferior results compared with those using a learned visual codebook~\cite{koh2023grounding,liu2023language}, or they require an exponential number of tokens to accurately represent an image~\cite{yu2023spae}. 
More importantly. relying on a frozen LLM means stopping adapting to follow human instructions through alignment with high-quality, instruction-following data~\cite{taori2023stanford,chiang2023vicuna}.
Hence, being trainable is a indispensable feature for MLLMs.

Considering this, we introduce \textbf{UniCode}, the first attempt to craft a \textbf{Uni}fied \textbf{Code}book for MLLMs, by integrating a VAE-style visual tokenizer~\cite{van2017neural} with the LLM.
To achieve this, we alternate the training process between these two modules, iteratively synchronizing the visual tokenizer's codebook with  the LLM's to maintain consistency.
This process, which we term "language-driven iterative training", utilizes a smooth moving average to udpate the visual codebook.
To further enhance the fidelity of images generated by UniCode, we introduce a novel pre-training task:  in-context image decompression. 
This task leverages in-context instructions to transform compressed image data into discrete visual tokens.
Furthermore, UniCode is designed to support stacked quantization~\cite{lee2022autoregressive,yu2023spae} to optimize visual tokenization efficiency, which compresses images into stacked code maps to reduce the feature resolution.
The unified codebook effectively converts visual inputs into language tokens.
Based on it, UniCode brodens the scope of visual instruction tuning ~\cite{liu2023language} by reformulating multimodal generation in the context of instruction-following format. 

Our key contributions can be summarized as follows:

\parskip=0.1em
\begin{itemize}[itemsep=0.1em,parsep=0em,topsep=0em,partopsep=0em]
    \item We propose UniCode, a innovative paradigm for MLLMs, featuring a unified codebook capable of tokenizing both visual and textual inputs. To achieve this, we adopt language-driven iterative training to learn such a codebook without additional parameters for visual-text alignment.
    \item We enrich the model's tuning with non-linguistic data integrated into the existing visual instructional dataset.  
    The tuning process is augmented by a unique in-context image decompression task, designed to improve the model's ability to interpret and generate complex multimodal content.
    \item Experimental analysis shows the effectiveness of UniCode compared to state-of-the-art MLLMs. Notably, this is achieved using a more efficient visual encoder that requires significantly fewer parameters and training samples.
\end{itemize}

%Meanwhile, the visual codebook is learned via smooth moving average.
%We term this procedure as language-driven iterative training.
%\vspace{-0.1cm}
\section{Related Work}
\label{sec:related}

\subsection{Visual Quantization}
Vector quantization (VQ) has achieved remarkable success in creating high-resolution images~\cite{van2017neural,vim,maskgit} and videos~\cite{maskvit,videogpt,teco}. 
VQ-VAE~\cite{van2017neural} firstly converts images into discrete representations and autoregressively models their distribution.
Following this work, Razavi~\etal\cite{razavi2019generating} adopt learned hierarchical representations, while Esser~\etal\cite{esser2021taming} introduce perceptual adversarial loss~\cite{wang2018perceptual} to refine the perceptual quality of reconstructed images.
Inspired by residual quantization~\cite{juang1982multiple,loshchilov2017decoupled}, Lee~\etal\cite{lee2022autoregressive} develop residual quantization (RQ), a technique that encodes images into a stacked map of discrete codes, thereby efficiently reducing the spatial resolution of features. 
You~\etal~\cite{you2022locally} propose hierarchical vector quantization (HQ) which employs a pyramid scheme with two-level codes for image encoding.
Despite these advancements, a limitation of these methods is that their codebooks, being jointly trained with the encoder and decoder, lack direct interpretability in natural language.
To address this, recent research has investigated to leverage frozen LLMs for image understanding~\cite{koh2023grounding,liu2023language}.
Liu~\etal\cite{liu2023language} innovate with LQAE, replacing the learned codebook with a text vocabulary from the frozen BERT~\cite{devlin2018bert}.
Despite its novelty, LQAE falls short in the fidelity of image reconstruction, underscoring the challenges of using a frozen LLM for content generation across modalities.
Yu~\etal\cite{yu2023spae} aim to solve the challenge by arranging quantized tokens in a multi-layer, coarse-to-fine pyramid.

\subsection{Multimodal Instruction Tuning}
In the field of natural language processing (NLP), previous studies have made significant strides in enabling LLMs~\cite{brown2020language,raffel2020exploring,zhang2022opt,chowdhery2022palm} to comprehend and execute natural language instructions, through a process known as instruction tuning~\cite{ouyang2022training}. 
Following this practice, recent efforts have extended its application to the multimodal realm~\cite{ye2023mplug,xu2022multiinstruct}.
Among these works, Liu~\etal\cite{liu2023visual} introduce LLaVA, the first model to apply the concept of visual instruction tuning to build a versatile visual assistant.  
Following this, Li~\etal\cite{li2023mimic} propose Mimic-it, enhancing the model's capability by incorporating multimodal in-context information directly into instruction data. 
Zhang~\etal\cite{zhang2023llavar} and Zhao~\etal\cite{zhao2023svit} have further research in this area by scaling instructional data and enriching it with text-dense images.
In addition to simply increasing data volume, Dai~\etal\cite{dai2023instructblip} develope InstructBLIP based on BLIP-2~\cite{li2023blip}, which introduces an advanced visual feature extraction mechanism to bolster performance across vision-language tasks.

While existing foundation models have marked impressive strides in multimodal benchmarks,  their capabilities are still limited to text-only generation.
Recent, a notable advancement Emu is introduced by Sun~\etal\cite{sun2023generative}, a model crafted for generative pretraining across multiple modalities. Despite its innovation, Emu necessitates a robust visual encoder with 1 billion parameters and relies on 80 million samples for effective pretraining.
Meanwhile, Lu~\etal\cite{lu2023unified} propose Unified-IO 2, a MLLM akin to our approach which encodes and generates text, vision, audio, and interleaved sequences.
Yet, it also requires significant computational demands with 1 billion image-text pairs.
Instead, UniCode diverges from the above approaches by focusing on learning a unified codebook.
We demonstrate through experiments that our model substantially decreases resource requirements while still achieving competitive results.

\section{Unified Codebook Learning}
\label{sec:method}

UniCode is built without bells and whistles for easy replication based on arbitrary transformer-based architecture of LLMs.
Our primary objective is to craft a unified codebook that efficiently tokenizes multimodal information.
To achieve this, we first give a brief overview of visual tokenization in Section~\ref{sec:vq}.
Then in Section~\ref{sec:uni_learn}, we propose our language-driven iterative training paradigm, after discussing various alternatives for synchronizing the learning of both visual and linguistic codebooks.
In Section~\ref{sec:uncomp_task}, we further propose a novel image decompression task designed for generation enhancement.

\subsection{Visual Tokenization}
\label{sec:vq}

Visual tokenization~\cite{van2017neural} is a process that compresses visual signals (e.g., images) into a series of discrete tokens, which generally consists of an encoder $\mathbbm{E}$, a decoder $\mathbbm{D}$ and a codebook $\mathbbm{C}=\{(k, e(k)) | k \in \{1,\cdots,{\rm K}\}\}$, where ${\rm K}$ denotes the codebook size.
Here, $\mathbbm{C}$ is a finite set of pairs, each consisting of a code $k$ and its corresponding $n$-dimensional code embedding $e(k) \in \mathbbm{R}^n$.
Similar to LLM, for each vector $z \in \mathbbm{R}^n$, the operation of visual tokenization $Q(z; \mathbbm{C})$ is defined to select the code from $\mathbbm{C}$ whose embedding is closest to $z$, which is described as:

\begin{equation}
   Q(z; \mathbbm{C})=\operatorname*{argmin}_{k\in \{1,\cdots,{\rm K}\}} \|z-e(k)\|_2^2.
\end{equation}

Given an image $\mathcal{I} \in \mathbbm{R}^{{\rm H} \times {\rm W} \times 3}$, the visual tokenizer first uses the encoder $\mathbbm{E}$ to derive its feature map $Z_0 \in \mathbbm{R}^{{ h} \times { w} \times c}$, with $c$ representing the embedding dimension.
Subsequently, each vector $z \in Z_0$ is assigned to the closest code within the codebook $\mathbbm{C}$, yielding a code map $M_{ij}=\mathcal{Q}({Z_0}_{ij}; \mathbbm{C})$ and its quantized feature map $Z_{ij}=e(M_{ij})$, where $i\in [1,h]$ and $j\in [1,w]$, respectively.
The decoder then utilizes $Z$ to reconstructs the image.
In this work, LLM is designed to either interpret these quantized embeddings as input, or to directly generate discrete tokens that can signify visual semantic concepts.

\noindent\textbf{Efficient Stack Quantization. }
As the resolution $(h \times w)$ of code map $M$ increases, the computational demand on LLM grows quadratically.
Given the LLM's inherent constraint on processing only a finite length of token sequences, reducing the resolution of the code map becomes crucial.
However, the fidelity of reconstruction is deeply influenced by the tokens' bit-depth~\cite{shannon1959coding}.
To strike a balance between efficiency and quality in visual tokenization, we consider stacked quantization as a viable solution~\cite{martinez2014stacked} to decrease the resolution of $M$.
Specifically, stacked quantization preserves the visual information by generating a $D$-layer code map $\hat{M}_d \in \mathbbm{N}^{\hat{h}_d \times \hat{w}_d \times D}$, where $d \in [1, D]$ and the dimensions $\hat{h}$, $\hat{w}$ are significantly reduced compared to $h$, $w$.
For each element $(i, j)$ in the code map, its ultimate embedding is an aggregation of $D$ quantized vectors $\hat{z}_{ij}=\mathcal{F}_{d=1}^D e(\hat{M}_{i,j,d})$, with $\mathcal{F}$ denoting the aggregation function (e.g., concatenation in HQ~\cite{you2022locally}, cumulative sum in RQ~\cite{lee2022autoregressive}).
In our study, HQ serves as a prime example for illustration.
Note that UniCode is adaptable to various variants of stacked quantization, making it a fertile area for further research.

In our approach, these visual tokens directly correspond to entries in the LLM's codebook, enabling our proposed UniCode to seamlessly interpret these aggregated quantized embeddings. 
Furthermore, we propose an image decompression task to enhance the LLM's capability for converting quantized embeddings into language tokens.

\subsection{Codebook Learning Paradigm}
\label{sec:uni_learn}

\begin{figure*}[!t]
\centering
\includegraphics[scale=0.45]{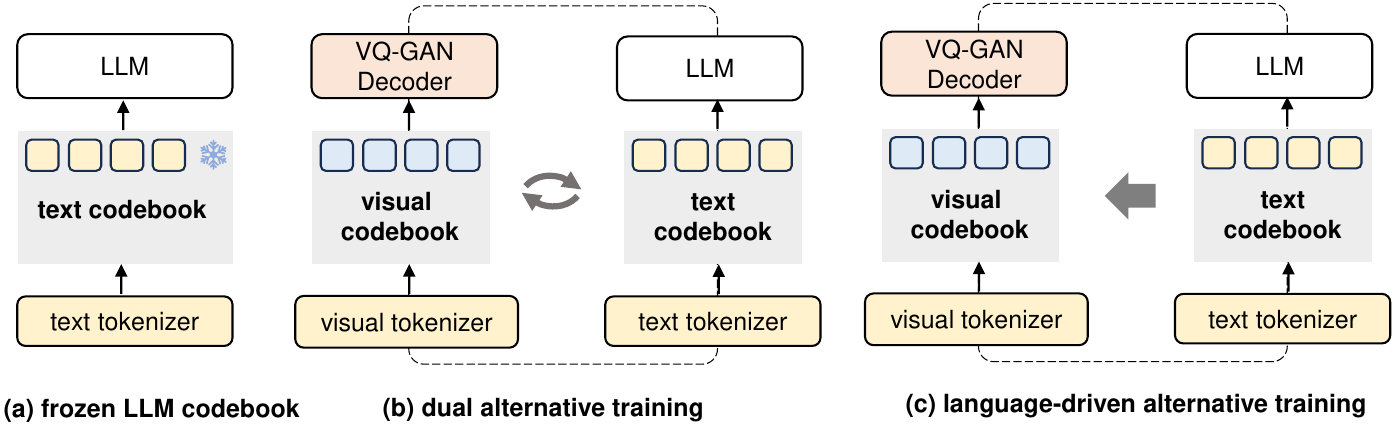}
\caption{
Illustration of multiple paradigms to obtain a unified codebook. Dotted line indicates the training loop:
\textbf{(a) frozen LLM codebook}, which initiates the codebook with a pretrained LLM and freezes it during training;
\textbf{(b) dual alternative training}, which jointly trains both visual tokenizer and LLM, by alternatively updating each one's codebook using the other's parameters.
\textbf{(c) language-driven iterative training}, which smoothly updates the codebook of visual tokenizer with LLM's through a moving average manner.}
\label{fig:method}
\end{figure*}

%As illustrated in Fig~\cite{}, we discuss multiple potential training strategy to sycnronize the codebook between vision and language:
Before introducing our proposed paradigm, we first discuss two alternatives to obtain a unified codebook:

\noindent\textbf{Frozen LLM Codebook. }
We start with a straightforward approach as illustrated in Figure~\ref{fig:method} (a), where the visual tokenizer's codebook is initialized with a pretrained LLM and simply keeps frozen during training.
While this approach directly links the visual tokenizer with the language vocabulary, it falls short in accurately capturing the semantic nuances in images.
Our empirical study further reveals that employing a frozen codebook adversely impacts the quality of reconstruction, especially for stacked quantization methods such as hierarchical quantization (HQ).
This can be primarily attributed to two factors: the absence of an explicit mechanism to synchronize the encoder/decoder with the frozen codebook, and the varying scales of multi-layer embeddings that divides the codebook into multiple parts~\cite{liu2023language}.
The pursuit of optimal reconstruction fidelity motivates the following development of dynamic alignment between the codebook and the encoder/decoder of visual tokenizer.

% \begin{minipage}[t]{0.32\textwidth}
% \centering
% \includegraphics[scale=0.45]{asserts/imgs/3.2.1.pdf}
% \caption{Frozen from Training}
% \end{minipage}
% \begin{minipage}[t]{0.32\textwidth}
% \centering
% \includegraphics[scale=0.4]{asserts/imgs/3.2.2.pdf}
% \caption{Dual Alternative Training}
% \end{minipage}
% \begin{minipage}[t]{0.32\textwidth}
% \centering
% \includegraphics[scale=0.4]{asserts/imgs/3.2.3.pdf}
% \caption{LLM-driven Iterative Training}
% \end{minipage}

\noindent\textbf{Dual Alternative Training. }
As shown in Figure~\ref{fig:method} (b), this approach dynamically aligns the visual tokenizer and LLM by alternating their training. 
In each training step of the visual tokenizer, its codebook is directly replaced by that of the LLM, and vice versa.
This approach ensures both modules are progressively optimized in a unified direction using a shared codebook.
However, a new challenge arises now from the disparity in the codebook change rate in the two modules. 
To be specific, the codebook change in the visual tokenizer is significantly greater than that of the LLM.
This becomes even more severe for stacked quantization due to their multi-layer code map, where each additional layer requires one more update of the codebook.
Such disparity finally leads to the misalignment between the codebook and LLM, resulting in the impairment of the LLM's language generation capabilities.

\noindent\textbf{Language-driven Iterative Training.}
To overcome the above issues and facilitate unified codebook learning, we introduce this paradigm as illustrated in Figure \ref{fig:method} (c).
Unlike dual alternative training, this approach does not employ the visual tokenizer to update the LLM's codebook.
Instead, we apply the exponential moving average (EMA) method~\cite{lee2022autoregressive} to ensure the codebook's alignment with the visual encoder, which dynamically updates the visual tokenizer’s codebook at a certain decay rate $\lambda$:

\begin{equation}
\small
\mathbbm{C}^{'} = \lambda \mathbbm{C} + (1-\lambda) \mathbbm{I} \cdot Z.
\label{eq:ema}
\end{equation}

$Z\in \mathbbm{R}^{hw \times c}$ represents the flattened features of a given image, as generated by the encoder. 
The indicator map $\mathbbm{I} \in \mathbbm{R}^{{\rm K} \times hw}$ summarises the usage of each code of $\mathbbm{C}$ in the feature map $Z$.
Crucially, at regular intervals, we integrate the codebook $\mathbbm{C}_L$ in the LLM to replace $\mathbbm{I}\cdot Z$ to update $\mathbbm{C}$, which can be denoted as:

\begin{equation}
\small
\mathbbm{C}^{'} = \lambda \mathbbm{C} + (1-\lambda) \mathbbm{C}_{L}.
\label{eq:ema2}
\end{equation}

Equation \ref{eq:ema2} ensures the gradual convergence of the visual tokenizer's codebook towards $\mathbbm{C}_L$ during training. 
Our paradigm not only aids in the efficient acquisition of a unified codebook, but also ensures the training of LLM remains undisturbed by the updates in the visual tokenizer.
Note that our paradigm is adaptable to various tuning approaches of LLM, including full parameter tuning, LoRA~\cite{hu2021lora}, or even freezing LLM.
%does not necessitate freezing the LLM.
%In fact, a key aspect of UniCode is tuning LLM to better respond to human instructions. 
%Various tuning methods, such as , can be employed in our LLM-driven iterative training process.
A significant distinction of our approach, compared to other MLLMs is that it does not need additional modules for visual-text alignment.
We believe this could be an alternative for unified MLLMs, especially considering the recent breakthrough of visual sequential modeling~\cite{bai2023sequential}.

\subsection{In-context Image Decompression}
\label{sec:uncomp_task}

\begin{wrapfigure}{r}{0.5\textwidth} 
\vspace{-0.3cm}
\centering
  \includegraphics[width=0.5\textwidth]{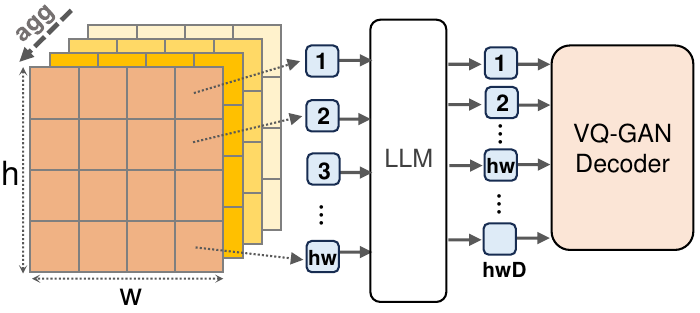}
  \vspace{-2mm}
  \caption{Illustration of the procedure for the in-context image decompression task, which accepts the compressed quantized embeddings $\hat{Z}\in \mathbbm{R}^{\hat{h}\times \hat{w}}$ as inputs, and then proceeds to transform these embeddings into their flattened codes $\hat{M}\in \mathbbm{R}^{\hat{h}\times \hat{w}\times D}$ that are subsequently used for visual decoding.
}
\label{fig:3.3} 
\vspace{-0.3cm}
\end{wrapfigure}

Since we adopt the stacked quantization as introduced in Section~\ref{sec:vq} to represent images with fewer tokens, UniCode encounters a misalignment issue when aggregating word embeddings with the LLM, which can hinder the learning of semantically meaningful tokens. 
To tackle this issue, we propose an image decompression pre-training task as shown in Figure~\ref{fig:3.3}, whose objective is to reconstruct the multi-layer code map $\hat{M}$ by feeding the LLM with the aggregated quantized embeddings $\hat{Z}$.
Initially, $\hat{Z}$ is processed into a flattened sequence of length $\hat{h}\times \hat{w}$.
We then define the target sequence as $\{u_1,u_2,...,u_{\hat{h}\times \hat{w} \times D}\}$, which is derived from $\hat{M}$, where each $u_{l} \in \hat{M}$.
Our goal is to maximize the likelihood of generation in an auto-regressive manner:

\begin{equation}
\small
\max\limits_{\theta} \sum_{l=1}^{\hat{h}\times \hat{w} \times D} \log P_{\Theta}(u_l|u_{<l}; \hat{Z}),
\end{equation}
where $\Theta$ denotes the trainable parameters of the LLM.
Moreover, to enhance our model’s capability in interpreting and generating across various modalities, we adopt a strategy similar to Liu~\etal\cite{liu2023visual}.
To be specific, we construct instruction-following pairs of multi-turn, conversation-style data for in-context learning: $\{\mathcal{X}_m^1, \mathcal{X}_z^1,..., \mathcal{X}_m^T, \mathcal{X}_z^T\}$.
Here, an image is segmented into $T$ pieces, with 
$\mathcal{X}_z^t$ and $\mathcal{X}_m^t$ representing the quantized embeddings and their corresponding visual codes for each segment $t$.
We organize these segments sequentially and consider each $\mathcal{X}_m^t$ as the response from the LLM.
For a sequence of length $L$, the probability of generating the target codes $\mathcal{X}_m$ is computed as:

\begin{equation}
\small
    P(\mathcal{X}_m^t|X_z^t) = \prod_{i=1}^{L} P_{\Theta}(x_i|\mathcal{X}_m^{<i}, \mathcal{X}_z^{<i}),
\end{equation}
where $\mathcal{X}_m^{<i}$ and $\mathcal{X}_z^{<i}$ denotes the visual codes and their compressed quantized embeddings in all segments before the current prediction token $x_i$, respectively.
We incorporate this task with our multimodal instruction tuning data, which mimics the misalignment between the compressed image embeddings and the LLM, encouraging our UniCode to generate images with higher quality.

\section{Training}

Following Liu~\etal\cite{liu2023improved}, we first leverage pairs of image-text data for multimodal instruction tuning.
Specifically, we organize each instructional instance into a sequence of multi-round dialogues, represented as $\{\mathcal{X}_q^1, \mathcal{X}_a^1,..., \mathcal{X}_q^N, \mathcal{X}_a^N\}$.
In this sequence, each pair $\{\mathcal{X}_q^i, \mathcal{X}_a^i\}$ signifies a question-answer round between a human and the chatbot assistant, with $N$ indicating the total number of dialogue rounds. 
This structured format is consistently applied throughout our instructional dataset.
%For text generation tasks, we directly utilize the instructional data proposed by LLaVA-1.5~\cite{liu2023improved}.
In addition, with the advancement that allows images to be represented as discrete language tokens, our model is capable of converting text-to-image samples (e.g., CC3M~\cite{sharma2018conceptual}) into instruction-answer pairs.
Lastly, we also prepare task-specific data using the same format for in-context image decompression as described in Section~\ref{sec:uncomp_task}.
We combine all the above data for multimodal instruction tuning.
To train our model efficiently, we adopt a negative log-likelihood objective over the prediction tokens:

\begin{equation}
\small
    \mathcal{L}(\Theta)=-\sum_{j=1}^{L} \log P_{\Theta}(y_j|\mathcal{I}, \hat{y}_{1:j-1}).
\end{equation}

Here, $y$ and $\hat{y}$ are used to represent the target and input token sequences, respectively, while $\Theta$ denotes the model parameters and $L$ denotes the length of the target sequence. 
Depending on the specific instruction provided, the input visual content, represented as $\mathcal{I}$, may correspond to an empty image. 
A notable aspect is the restriction of loss computation exclusively to the answer tokens $\mathcal{X}_a$, which is designed to avoid oversimplifying the training process and to ensure that the model remains focused on generating accurate and coherent responses.
We adopt a two-stage instruction-tuning process to train UniCode.
It is important to note that our training process does not include the multimodal alignment stage, which is different from Liu~\etal\cite{liu2023visual}.

\noindent\textbf{Stage I: Unified Codebook Learning. } 
Our goal in this stage is to align the visual tokenizer with the LLM to share one codebook.
We train the visual tokenizer through the image reconstruction task. 
There is no limitation on the type of training images, and in practice, more diverse and large-scale data can bring better performance to the visual tokenizer. 
To strike a balance between performance and efficiency, we only consider a limited scale in this work.
For the LLM, it requires textual instruction-answer data to enhance its ability to follow instructions~\cite{chiang2023vicuna}.
We alternate training process between these two modules, updating the codebook parameters using our language-driven iterative paradigm.
After Stage 1, UniCode obtains a unified codebook that can simultaneously represent non-linguistic signals to achieve multimodal Input/Output (I/O).

\noindent\textbf{Stage II: Multimodal Instruction Tuning. } 
In the second stage, we keep the visual encoder and decoder frozen while exclusively fine-tuning the LLM.
This stage fully utilizes the comprehensive multimodal instructional dataset, which augments the model's effectiveness in interpreting and responding to intricate multimodal instructions.
The focus of this stage is on the model's ability to produce multimodal outputs, thereby significantly enriching its multimodal comprehension and response capabilities.

%, especially stacked quantization, undergoes more frequent and significant updates in its codebook compared to the LLM.
%is designed to synergistically optimize both modules towards the same direction using a shared codebook.
%However, new challenge arises from the disparity in codebook change between the two modules: the change of the visual tokenizer's codebook is significantly greater than that of the LLM's codebook, especially for stacked quantization methods, which, due to their multi-layer structure with more frequent updates of the codebook.

%To ensure optimal reconstruction fidelity, this motivates the development of dynamic alignment between the codebook and the encoder/decoder of visual tokenizer next.

%As these visual tokens directly correspond to entries in the LLM's codebook, our proposed UniCode can more readily interpret these aggregated quantized embeddings. 
%We leave the exploration of more stacked quantization approaches in the further.
%However, the reconstruction quality depends on the number of bits that tokens contain~\cite{shannon1959coding}.
%Further exploration is a promising avenue for future research.

%While we focus on HQ in this work, 
\section{Experiments}
\label{sec:experiments}
To throughly evaluate the expansive multimodal capabilities of UniCode, 
we first conduct a series of ablation studies in Section~\ref{sec:abl}.
We then carry out comparison experiments across several key benchmarks: image generation (Section~\ref{sec:gen}), image reconstruction (Section~\ref{sec:recon}), and multimodal understanding (Section~\ref{sec:vqa}).
Due to the space limitation, more details, visualization, and experimental results can be seen in our appendix.

\subsection{Implemented Details}
During Stage I, we train the visual tokenizer on the LCS-558K dataset introduced by LLaVA~\cite{liu2023visual}.
We directly employ a pretrained LLM~\cite{chiang2023vicuna}, opting not to conduct further instruction tuning on text-only data.
It's worth mentioning that this stage  is designed with flexibility to extend and allow the LLM to undergo pretraining on a extensive text corpora with full parameter tuning.
In Stage II, we focus on fine-tuning the LLM using a curated combined dataset, which includes Mixed-665K~\cite{liu2023visual}, the text-to-image dataset CC3M ~\cite{sharma2018conceptual}, and our specially tailored data for the in-context image decompression task.

\begin{table}[h]
\centering
\vspace{-5mm}
\caption{Comparisons of different paradigms for MLLMs on VQA and image generation benchmarks. Here, "tok" is used as an abbreviation for "tokenizer."}
\setlength{\tabcolsep}{5pt}
\scalebox{0.8}{
\begin{tabular}{c|l|l|l|l|l|l|l|l}
\toprule
\multirow{2}{*}{paradigm} & \multicolumn{5}{c}{VQA Benchmarks} & \multicolumn{3}{|c}{Image Gen (FID $\downarrow$)} \\
\cline{2-9}
%\midrule
  & VQA$^2$ & VizWiz & SQA & VQA$^{\rm T}$ & POPE & ImgNet & LSUN-cat & LSUN-church  \\
\midrule
vis enc+text tok  & 52.3 & 45.4 & 62.2 & 42.1 & 69.7 & -  &  -  & - \\
vis tok+text tok  & 49.0 & 44.5 & 56.7 & 37.8  & 65.4 & 9.82 & 10.28 & 10.78   \\
unified tok  & 53.1 & 46.2 & 62.9 &  42.5 & 71.8 & 6.72 & 8.07 & 6.96  \\
\bottomrule
\end{tabular}}
\vspace{-8mm}
\label{tab:abl_mllm}
\end{table}

\subsection{Ablation Study}
\label{sec:abl}

\noindent\textbf{Comparison of different paradigms for MLLMs.}
In Table~\ref{tab:abl_mllm}, we compare three MLLM paradigms as depicted in Figure~\ref{fig:intro}.
Notably, the use of a unified codebook (Row 3) obtains stable improvements in both VQA and image generation benchmarks compared to the separate use of visual and text tokenizers (Row 2).
This can be attributed to the aligned distribution of shared tokens and pretrained LLM, which also results in greater resource efficiency during both training and inference.
Additionally, the unified codebook also exhibits slight improvement in VQA tasks compared with using ``visual encoder+text tokenizer" (Row 1).
Note that Row 1 lacks direct applicability to image generation, instead, our unified codebook enables LLMs to produce multimodal outputs.

\begin{wraptable}{r}{0.58\textwidth}
\centering
\setlength{\tabcolsep}{3pt}
\vspace{-8mm}
\caption{Comparisons of different visual encoder setups, where ``cc3m imgs'' and ``GT imgs'' refer to using additional images in CC3M~\cite{sharma2018conceptual} and evaluation groundtruth to train the visual encoder, ``w/ ViT$^*$'' denotes using the pretrained and larger ViT encoder~\cite{fang2023eva} instead of training it from scratch.}
\scalebox{0.8}{
\begin{tabular}{l|l|l|l|l|l}
\toprule
\multirow{2}{*}{Setup} & \multicolumn{5}{c}{VQA Benchmarks}  \\
\cline{2-6}
  & VQA$^2$ & VizWiz & SQA & VQA$^{\rm T}$ & POPE  \\
\midrule
UniCode  & 53.1 & 46.2 & 62.9 &  42.5 & 71.8   \\
+ cc3m imgs  & 53.6 & 47.4 & 64.3 & 45.6 &  74.3    \\
++ GT imgs  & 53.7 & 47.4 & 64.8 &  44.9 & 75.1   \\
+++ w/ ViT$^*$ & 56.2 & 47.1 & 65.4 & 47.3 & 77.6   \\
\bottomrule
\end{tabular}
}
\vspace{-6mm}
\label{tab:abl_vis_enc}
\end{wraptable}

\noindent\textbf{Comparison of different visual setups. }
UniCode employs a relatively lightweight visual encoder and is trained on a modest dataset of 558K images~\cite{liu2023improved}.
While efficient, this setup restricts its ability to extract comprehensive visual features and hinders its generalization to novel contexts. 
This limitation becomes more evident when compared to models like CLIP~\cite{radford2021learning}, which benefits from training on a vast collection of 400 million image-text pairs.
In Table~\ref{tab:abl_vis_enc}, we demonstrate that the limitation can be mitigated.
By enriching the dataset with additional images from CC3M (Row 2) and evaluation groundtruth (Row 3), UniCode manifests consistant  improvements across various VQA benchmarks.
Additional improvement can be obtained by replacing the visual encoder trained from scratch with a pretrained and advanced version (Row 4).

%\begin{table}[h]
\begin{wraptable}{r}{0.58\textwidth}
\centering
\vspace{-8mm}
\setlength{\tabcolsep}{4pt}
\caption{Comparison of different visual tokenizers. Better tokenizer brings better performance.}
\scalebox{0.8}{
\begin{tabular}{l|l|l|l|l|l}
\toprule
 & VQA$^{2}$ & VizWiz & SQA & VQA$^{T}$ & POPE    \\
\midrule
VQ-GAN~\cite{esser2021taming}        & 49.1 & 42.6 & 60.8 & 41.2 & 65.1  \\
RQ-VAE~\cite{lee2022autoregressive}  & 49.8 & 44.0 & 61.5 & 41.6 & 67.5  \\
HQ-VAE~\cite{you2022locally}         & 53.1 & 46.2 & 62.9 & 42.5 & 71.8 
\\
\bottomrule
\end{tabular}}
\vspace{-6mm}
\label{tab:abl_vis_tok}
%\end{table}
\end{wraptable}
%Results on five additional text-image datasets: QGA, MME, MMBench (MMB), MMBench-CN and SEED-Bench. Here $*$ denotes using cc3m

In addition to the visual encoder, we also verify the influence of different visual tokenizers as demonstrated in Table~\ref{tab:abl_vis_tok}.
It is crucial to note again that UniCode is designed to be compatible with a wide range of visual quantization approaches. 
Furthermore, we observe that as we keep upgrading the visual tokenizer (from Row 1 to Row 3), the performance of UniCode is also improved.
Such observations confirm that: the UniCode's overall capabilities can be continuously enhanced by keeping enhancing the visual setup.

\begin{table}[h]
\centering
\vspace{-5mm}
\caption{Comparisons of different paradigms to learn a unified codebook.}
\setlength{\tabcolsep}{4pt}
\scalebox{0.8}{
\begin{tabular}{c|l|l|l|l|l|l|l|l}
\toprule
\multirow{2}{*}{paradigm} & \multicolumn{5}{c}{VQA Benchmarks} & \multicolumn{3}{|c}{Image Gen (FID $\downarrow$)} \\
\cline{2-9}
%\midrule
  & VQA$^2$ & VizWiz & SQA & VQA$^{\rm T}$ & POPE & ImgNet & LSUN-cat & LSUN-church  \\
\midrule
frozen  & 44.2 & 35.1 & 56.8 & 36.3 & 63.9 & 34.45 &  33.84  & 34.26 \\
dual  & 9.3 & 5.2 & 11.2 & 8.5  & 13.2 & 8.87 & 9.76 & 9.54   \\
iter  & 53.1 & 46.2 & 62.9 &  42.5 &  71.8 & 6.72 & 8.07 & 6.96   \\
\bottomrule
\end{tabular}}
\vspace{-5mm}
\label{tab:abl_uni}
\end{table}

\noindent\textbf{Comparison of different paradigms to learn the unified codebook. }
We include the relevant results in Table~\ref{tab:abl_uni}.
The dual alternative training (dual) results in a performance collapse, particularly in VQA benchmarks.
This issue arises from a disruption in the consistency between the LLM architecture and codebook, as discussed in Section~\ref{sec:uni_learn}.
In addition, our paradigm (iter) brings representative visual tokens, therefore leading to notable improvements in visual generation compared with the frozen LLM codebook (frozen).

\begin{wraptable}{r}{0.48\textwidth}
%\begin{table}[h]
\centering
\vspace{-1cm}
\setlength{\tabcolsep}{4pt}
\caption{Ablation of in-context image decompression task (``ImgDe'') on image generation tasks. We use FID as the metric.}
\scalebox{0.8}{
\begin{tabular}{llll}
\toprule
Method      & ImageNet  & CC3M & LSUN-Cat \\ 
\midrule
w/o ImgDe    & 7.08    & 11.91 & 8.53  \\
w/ ImgDe     & 6.72    & 11.54 & 8.07  \\
\bottomrule
\end{tabular}}
\vspace{-6mm}
\label{tab:abl_img_decomp}
%\end{table}
\end{wraptable}

\noindent\textbf{Effect of in-context image decompression task. }
The results in Table~\ref{tab:abl_img_decomp} demonstrate that our pretraining task clearly enhances the visual generation quality of our model across various configurations: class-conditioned (ImageNet), text-conditioned (CC3M), and unconditioned (LSUN-Cat).
We posit that this enhancement in performance can be attributed to the pretraining task's ability to prevent premature convergence.
It achieves this by escalating the complexity of the training process and enriching the diversity of the training samples.

%\begin{table}[h]
\begin{wraptable}{r}{0.45\textwidth}
\centering
\vspace{-3mm}
\caption{Ablation of different code map resolutions on VQA benchmarks.}
\setlength{\tabcolsep}{3.5pt}
\scalebox{0.8}{
\begin{tabular}{l|lllll|l}
\toprule
 & 192 & 256 & 320 & 384  & Raw &  320$^{*}$   \\ 
\midrule
VQA$^{\rm 2}$ & 38.2 & 53.1 & 41.3 & 42.6 & 36.4 & 54.5  \\
VizWiz        & 42.8 & 46.2 & 43.9 & 41.3 & 39.1 & 47.1 \\
SQA$^{\rm I}$ & 61.1 & 62.9 & 63.7 & 62.0 & 59.2 & 63.8  \\
\bottomrule
\end{tabular}}
\vspace{-7mm}
\label{tab:abl_code_res}
%\end{table}
\end{wraptable}

\noindent\textbf{Effect of different code map resolution. }
In Table~\ref{tab:abl_code_res}.
spanning Column 1-5, UniCode is pretrained with images of resolution 256$\times$256, and reaches optimal performance when tested at this identical resolution.
Notably, there is a marked decrease in performance when test resolutions are increased beyond this point, even though these larger resolutions do not exceed the LLM's token length capacity.
We  deduce that this drop in performance stems from a misalignment between training and testing conditions.
Specifically, testing with resolutions significantly larger than those used in training creates a disparity in how each element of the code map represents image areas.
In Column 6, we verify this hypothesis by pretraining UniCode using 320$\times$320 images (320$^*$), and the results of our model are improveed to our expectation.

\begin{minipage}[t]{0.45\textwidth}
\centering
\setlength{\tabcolsep}{3pt}
\captionof{table}{Comparison of FIDs for unconditioned image generation on LSUN-\{Cat, Bedroom, Church\}~\cite{yu2015lsun}.}
\scalebox{0.8}{
\begin{tabular}{llll}
\toprule
\multirow{2}{*}{Method} & \multicolumn{3}{c}{FID $\downarrow$} \\  
 & Cat   & Bedroom & Church \\ 
\midrule
ImageBART      & 15.09 & 4.90    & 7.89   \\
StyleGAN2~\cite{karras2020analyzing}      & 7.25  & 2.35    & 3.86   \\
VQ-GAN         & 17.31 & 6.35    & 7.81   \\
RQ-Transformer & 8.64  & 3.04    & 7.45   \\ \midrule
%UniCode w/o ImgDe      & 8.69  & 3.22    & 7.53  \\
%UniCode w/o CC3M       & 8.47  & 2.97    & 7.34  \\
HQ-TVAE$^*$              & 8.35  & 2.89    & 7.12  \\
HQ-UniCode               & 8.07  & 2.65    & 6.96  \\
\bottomrule
\end{tabular}}
\label{tab:comp_uncon_gen}
\end{minipage}
\hfill
\begin{minipage}[t]{0.45\textwidth}
\centering
\setlength{\tabcolsep}{3pt}
\captionof{table}{Comparison of FIDs and CLIP score for text-conditioned image generation on CC3M validation set.}
\scalebox{0.8}{
\begin{tabular}{llll}
\toprule
Method    & Params   & FID $\downarrow$ & CLIP-s $\uparrow$ \\ 
\midrule

ImageBART    & 2.8B & 22.61   & 0.23  \\
LDM~\cite{rombach2022high}  & 645M & 17.01 & 0.24 \\
\midrule
VQ-GAN         & 1.5B & 28.86 & 0.20 \\
RQ-Transformer & 654M & 12.33 & 0.26 \\
HQ-TVAE        & 579M & 12.86    & 0.26  \\
\midrule
HQ-TVAE$^*$    & 7B   & 12.13    & 0.28  \\
HQ-UniCode     & 7B   & 11.54    & 0.30  \\
\bottomrule
\end{tabular}}
\label{tab:comp_text_gen}
\end{minipage}

\subsection{Comparison on Image Generation}
\label{sec:gen}

\begin{wraptable}{r}{0.48\textwidth}
%\begin{table}[h]
\centering
\vspace{-1.2cm}
\setlength{\tabcolsep}{4pt}
\caption{Comparisons of FIDs and ISs for class-conditioned image generation on ImageNet~\cite{deng2009imagenet}.}
\scalebox{0.8}{
\begin{tabular}{lllll}
\toprule
Method    & Params   & FID $\downarrow$ & IS  $\uparrow$ \\ 
\midrule
ADM~\cite{dhariwal2021diffusion}  & 554M & 10.94    & 101.0   \\
ImageBART~\cite{esser2021imagebart}    & 3.5B & 21.19    & 61.6  \\
VQ-Diffusion~\cite{gu2022vector}  & 370M & 11.89 & \\
\midrule
VQ-VAE-2~\cite{razavi2019generating}     & 13.5B & $\approx$ 31 & $\approx$ 45   \\
VQ-GAN~\cite{esser2021taming}          & 1.4B & 15.78 & 74.3   \\
RQ-Transformer~\cite{lee2022autoregressive}      & 3.8B  & 7.55 &  134.0   \\ 
HQ-TVAE~\cite{you2022locally}      & 1.4B  & 7.15 & -   \\
\midrule
HQ-TVAE$^*$    & 7B  & 7.04 & 171.4   \\
HQ-UniCode    & 7B  & 6.72    & 208.9  \\
\bottomrule
\end{tabular}}
\vspace{-8mm}
\label{tab:comp_gen_class}
%\end{table}
\end{wraptable}

%\midrule
%BigGAN~\cite{brock2018large} & 164M & 7.53     & 168.6 \\
%BigGAN-deep~\cite{brock2018large} & 112M & 6.84 &  203.6 \\
%DCT~\cite{nash2021generating}    & 738M & 36.5 & n/a   \\
%w/o ImgDe &    & 7B  & 15.48    & 89.3  \\

\begin{figure}[t]
\centering
\includegraphics[width=0.85\textwidth]{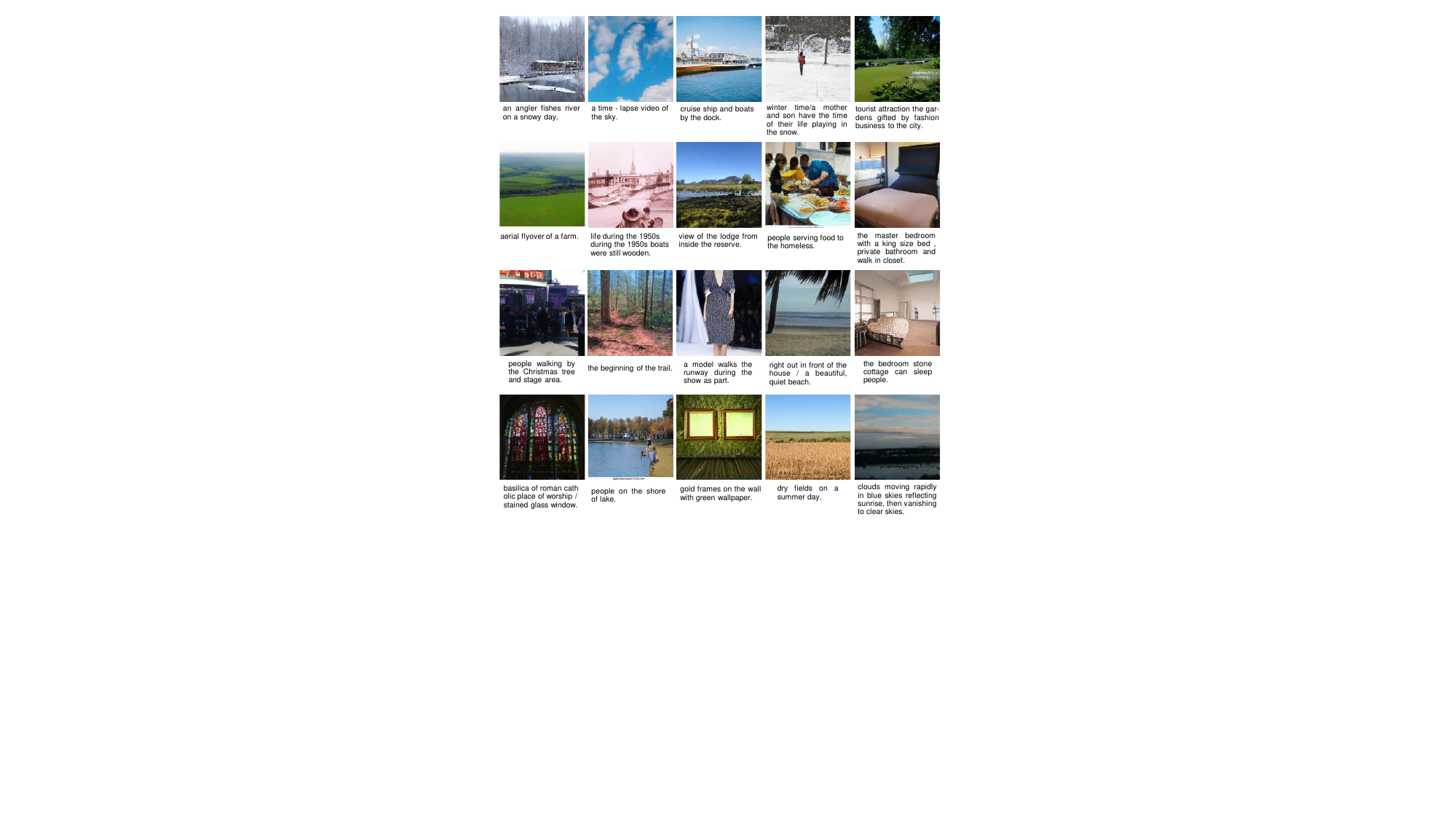}
\caption{
Qualitative examples of text-conditioned image generation on CC3M.
}
\label{fig:cc3m_gen} 
%\vspace{-0.5cm}
\end{figure}

We first assess the capability of our model in unconditioned image generation in Table~\ref{tab:comp_uncon_gen}, utilizing three subsets of the LSUN dataset.
Initially, we combine ImageNet with the LCS-558K dataset to pretrain our visual tokenizer, then finetune the model for another one epoch on the downstream dataset.
Given the extensive size of the dataset, we opt for LoRA~\cite{hu2106low} to finetune LLM to avoid overfitting.
Due to the lack of training details for HQ-TVAE, we have implemented its 7B version (HQ-TVAE$^*$) and compare it with our model (HQ-UniCode) for a fair comparison based on the same parameter setup.
Our model performs clearly better than HQ-TVAE$^*$.
Furthermore, we carry out experiments on text-conditioned image generation in Table~\ref{tab:comp_text_gen} and class-conditioned generation in Table~\ref{tab:comp_gen_class}, UniCode obtains similar improvements on these two benchmarks.
%We then compare with HQ-TVAE based on the same parameter number for a fair comparison.
As can be seen, the improved results demonstrate the benefit of employing a unified codebook to enhance visual generation, especially considering that our reconstruction quality is suboptimal compared with original HQ as discussed in Section~\ref{sec:recon}. 
We attribute this benefit to the alignment between the unified codebook and LLM’s textual space.
Lastly, we present some qualitative examples as shown in Figure~\ref{fig:cc3m_gen}. 
More visualization cases can be seen in our appendix.

\subsection{Comparison on Image Reconstruction}
\label{sec:recon}

%\begin{table}[h]
\begin{wraptable}{r}{0.51\textwidth}
\centering
\vspace{-1.2cm}
\caption{Comparison of reconstruction quality on ImageNet and LCS-558K datasets, according to their codebook size ($\rm K$), resolution (Res), number of used layers and tokens.}
\scalebox{0.75}{
\begin{tabular}{l|ccc|cc}
\toprule
\multirow{2}{*}{Tokenizer} & \multirow{2}{*}{Res} & \multirow{2}{*}{\begin{tabular}[c]{@{}l@{}}Layers:\\ Tokens\end{tabular}} & \multirow{2}{*}{K}  & \multicolumn{2}{c}{rFID $\downarrow$} \\ \cline{5-6} 
 & &  & & Imagenet  & LCS-558K \\ 
\midrule
VQ-GAN & 64 & 1:64 & 16384 & 17.95 & 23.83 \\
VQ-GAN & 256 & 1:256 & 16384 & 4.9 & 11.26  \\
SPAE~\cite{yu2023spae} & 256 & 5:341 & 16384 & 9.49 & - \\
SPAE & 256 & 6:597 & 16384 & 4.41 & - \\
%RQ-VAE & 64 & 4:256  & 16384 & 4.73 & 10.97 \\
RQ-VAE & 64 & 4:256 & 32000 & 6.82 & 12.09 \\
HQ-VAE & 256 & 2:320 & 32000 & 2.61 & 8.35 \\
\midrule
%RQ-frozen & 64 & 8:512  & 32000 & 331.22 & 374.45 \\
%RQ-dual & 64 & 8:512  & 32000 & 3.71 & 9.23 \\
RQ-UniCode & 64 & 8:512 & 32000 & 3.78 & 9.33 \\
HQ-UniCode & 256 & 2:320 & 32000& 2.83 & 7.91     \\
\bottomrule
\end{tabular}}
\vspace{-3mm}
\label{tab:img_recon}
%\end{table}
\end{wraptable}

\begin{wrapfigure}{*}{0.45\textwidth} 
\vspace{-0.6cm}
\centering
\includegraphics[width=0.45\textwidth]{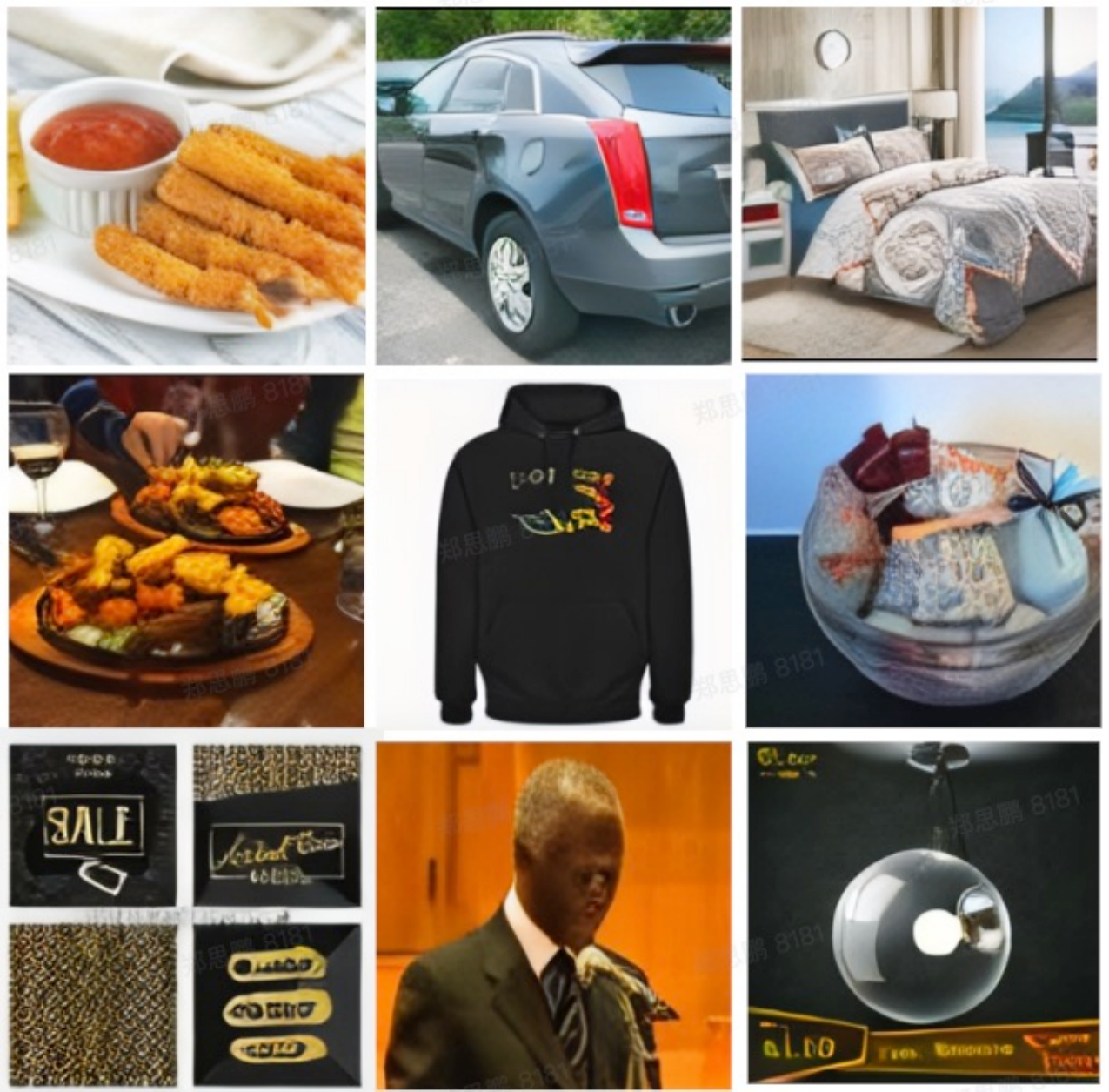}
\caption{Qualitative examples of image reconstruction generated by our proposed UniCode. Their raw images can be seen in the appendix.
%We observe that although the VAE-style visual tokenizer can roughly depict the scenes in images, it still falls short in capturing fined-grained details (such as text and faces in images), especially in datasets with rich scenes like CC-SBU, when compared to other visual models.
}
\label{fig:recon} 
%\end{figure}
\vspace{-0.6cm}
\end{wrapfigure}

Table~\ref{tab:img_recon} validates the reconstruction quality of the visual tokenizer of our model.
Such validation is crucial to ensure that the tokenizer preserves essential semantics after visual quantization.
In this table, we observe that multi-layer stacking, as a form of stacked quantization structure, substantially boosts the model's ability to efficiently represent images.
However, this benefit comes with a trade-off: an increased number of layers significantly lengthens the sequence, posing a greater challenge for decoding by LLM.
In comparison with HQ or RQ, the reconstruction quality of UniCode, when using the unified codebook, is nearly on par, indicating that our learning paradigm for the unified codebook does not significantly damage VAE training.
In Figure~\ref{fig:recon}, we present qualitative examples of image reconstruction on LCS-558K~\cite{liu2023improved}. 
When compared to ImageNet, there is a significant decline in reconstruction quality on LCS-558K, probably attributed to LCS-558K's more diverse scenes.

% Use figure* for multi-column figure

\begin{table}[t]
\centering
\caption{
Comparison with MLLMs on VQA benchmarks. 
UniCode outperforms another multimodal generation model Emu.
It achieves competitive results against other methods while requiring less data and fewer parameters for its visual tokenizer.
Here, ``M2T'' and ``M2M'' refer to the model's capability to generate either text only or multiple modalities.
``Vis-P'', ``PT'' and ``IT'' represent the number of parameters in the visual encoder, the number of samples for multimodal alignment, and instruction tuning, respectively.
Results on more benchmarks are provided in the appendix.
}
\scalebox{0.79}{
\begin{tabular}{l|l|llll|lllllll}
\toprule
Method   &  Type  & LLM & Vis-P
& PT & IT & VQA$^{\rm v2}$ & VizWiz & SQA$^{\rm I}$ & VQA$^{\rm T}$ & POPE & MMB & MMB$^{\rm CN}$ \\
\midrule
BLIP-2~\cite{li2023blip} & M2T & Vicuna-13B & 303M & 129M & 0 & 41.0 & 19.6 & 61 & 42.5 & 85.3 & - & - \\
InstructBLIP~\cite{dai2023instructblip} & M2T & Vicuna-7B  & 303M & 129M & 1.2M & - & 34.5 & 60.5 & 50.1 & - & 36 & 23.7 \\
Qwen-VL~\cite{bai2023qwen} & M2T    & Qwen-7B  & 1.8B  & 1.4B & 50M & 78.8 & 35.2 & 67.1 & 63.8 & - & 38.2 & \\
Emu~\cite{sun2023generative} & M2M  &  LLaMA-13B   & 1B & 82M & 240K & 52.0  &  34.2  & - & - & - & - & - \\
Emu-I~\cite{sun2023generative}  & M2M  &  LLaMA-13B   & 1B & 82M & 240K & 40.0  &  35.4  & - & - & - & - & -\\
LLaVA-1.5 & M2T  & Vicuna-7B  & 303M  & 558K & 665K & 79.1 &  47.8 & 68.4 & 58.2 & 86.4 & 64.3 & 58.3\\
\midrule
UniCode & M2M &  Vicuna-7B & 104M & 0 & 665K & 53.1 & 46.2 & 62.9 & 42.5 & 71.8 & 33.7 & 25.5 \\
UniCode+ & M2M &  Vicuna-7B & 1B & 0 & 665K & 56.2 & 47.1 & 65.4 & 47.3 & 77.6 & 37.2 & 29.1 \\
\bottomrule
\end{tabular}
}
\label{tab:comp_vqa}
\end{table}

%The symbol $^*$ denotes the chat version of LLaMA-2.
%of reconstruction quality on the ImageNet validation and CC3M$^\dagger$ sets, according to their codebook size ($K$), resolution (Res), number of used layers and tokens. 
%Here CC3M$^\dagger$ denotes the 558K subset in LLaVA~\cite{liu2023visual}

%IDEFICS-80B~\cite{idefics2023}  & M2T  & LLaMA-65B &  632M & - & 353M & 1M & 60.0 &  36.0 & - & 30.9  & - \\
%Qwen-VL-Chat~\cite{bai2023qwen} & M2T & Qwen-7B  & 1.8B  & 256 & 1.4B & 50M & 78.2 & 38.9 &  68.2 & 61.5 & - \\
%\textcolor{blue}{UniCode} & M2M   & LLaMA-2-7B$^*$ & 104M & 169 & 0 & 665K & 52.3 &   35.9 & 63.7 & 44.8 & 65.3 \\
%LLaVA-1.5~\cite{liu2023improved} & M2T & LLaMA-2-7B$^*$ & 303M & 336 & 558K & 665K & 76.6 &  41.9 & 65.7 & 52.4 & 86.1 \\

\subsection{Comparison on Multimodal Understanding}
\label{sec:vqa}

% Use figure* for multi-column figure
%\begin{figure}[tp]

We first carry out experiments on a diverse set of seven benchmarks in Table~\ref{tab:comp_vqa}, including VQA-v2 (VQA$^{\rm v2}$)~\cite{goyal2017making}, VizWiz~\cite{gurari2018vizwiz}, ScienceQA-IMG (SQA$^{\rm I}$)~\cite{lu2022learn}, TextVQA (VQA$^{\rm T}$)~\cite{singh2019towards}, POPE~\cite{li2023evaluating}, MMB~\cite{liu2023mmbench} and MMB$^{\rm CN}$.
Experimental results on more benchmarks are provided in our appendix.
It is encouraging that our model performs considerably well even with the smallest scale of training data and fewer parameters.
UniCode outperforms many recently proposed MLLMs in several benchmarks. More importantly, it obtains stable improvement on both VQA$^{\rm v2}$ and VizWiz benchmarks when compared to another multimodal generation model Emu~\cite{sun2023generative}.
Through these experiments, we validate the feasibility of a unified codebook as an alternative paradigm for multimodal generative models.
When compared to the current state-of-the-art model LLaVA-1.5, UniCode shows significant performance variations across different benchmarks. 
For example, UniCode's performance is competitive with LLaVA-1.5 in the VQA$^{\rm T}$ and SQA$^{\rm I}$ benchmarks.
However, it lags significantly (nearly 20\%) behind in the POPE~\cite{li2023evaluating} benchmark.
We speculate that this is likely due to the insufficient training data provided for the visual tokenizer, which leads to the limitation of the tokenizer in terms of generalization.

UniCode initially employs a lightweight visual tokenizer, which, due to limitations in resolution, training data, and the scale of parameters, results in suboptimal performance.
To address these shortcomings, we have developed an enhanced variant, referred to as 'UniCode+'. 
UniCode+ incorporates a more substantial dataset for training and integrating a pretrained and larger ViT encoder as detailed in Table~\ref{tab:abl_vis_enc}.
As demonstrated in Table~\ref{tab:comp_vqa}, UniCode+ significantly outperforms the original UniCode across all VQA benchmarks. 
This improvement underscores the potential for elevating model performance through the adoption of a more sophisticated visual encoder.

\section{Conclusion}
\label{sec:conclusion}
We introduce UniCode, a pioneering effort in the Multimodal Language Learning Model (MLLM) field to create a unified codebook for both visual and textual tokenization.
UniCode innovates with a language-driven iterative training paradigm and an in-context image decompression task, enabling the unified codebook to facilitate multimodal instruction tuning for non-linguistic generation tasks. 
Our comprehensive experiments in multimodal understanding and generation, coupled with an extensive ablation study, position UniCode as a promising new approach for advancing research within the MLLM community.

%We propose UniCode, the first attempt of MLLM to learn a unified codebook for visual and text tokenization.
%By proposing the language-driven iterative training paradigm and in-context image decompression task for codebook learning, UniCode can employ the unified codebook to expand multimodal instruction tuning towards non-linguistic generation. 
%Through experiments on multimodal understanding and generation and extensive ablation analysis, we believe that UniCode offers a potentially new paradigm for the MLLM community.
%which can potentially open up new directions for future research in multimodal foundation models.

\bibliographystyle{splncs04}
\bibliography{reference}
\end{document}